\documentclass[runningheads]{llncs}

\usepackage{eccv}
\usepackage{eccvabbrv}
\usepackage{graphicx}
\usepackage{booktabs}
\usepackage{tabularx}
\usepackage{array}
\usepackage{amsmath,amssymb}
\usepackage{xcolor}
\usepackage{microtype}
\ifdefined\XeTeXversion
  \newcount\pdfcompresslevel
  \newcount\pdfoptionpdfminorversion
  \newcount\pdfgentounicode
  \def\pdfglyphtounicode#1#2{}
\fi
\usepackage[accsupp]{axessibility}
\usepackage[hidelinks]{hyperref}

\newcommand{\J}{\mathcal{J}}
\newcommand{\F}{\mathcal{F}}
\newcolumntype{Y}{>{\centering\arraybackslash}X}

\begin{document}
\raggedbottom

\title{Agreement-Based Audio-Visual Segmentation: Champion Report for the MeViS-Audio Track in the 8th LSVOS Challenge}
\titlerunning{Agreement-Based Audio-Visual Segmentation}

\author{Yiwen Ren\inst{1} \and
Jianing Liu\inst{1} \and
Yingxin Wang\inst{1} \and
Kexin Zhang\inst{1} \and
Licheng Jiao\inst{1} \and
Lingling Li\inst{1} \and
Xu Liu\inst{1}}
\authorrunning{Y. Ren et al.}
\institute{National 111 Project Base of Intelligent Information Processing}

\maketitle

\begin{abstract}
The MeViS-Audio track asks a system to segment the objects described by a
spoken motion expression throughout a video and to return empty masks when the
described target is absent.  We present a simple staged solution.  Qwen3-ASR
first converts speech into text.  Several video mask tracks are then produced
with complementary grounding and segmentation models.  Instead of trusting a
single prediction, we select the track that has the highest average mask
agreement with the other candidates.  A small set of explicit direction,
count, and plural rules corrects queries that require more than ordinary
single-object tracking.  Finally, a video-level classifier combines visual,
audio-visual, and within-video query scores to decide whether any target is
present.  The submitted system obtains 0.5952 $\J\&\F$, 0.7931 no-target
accuracy, 0.9205 target accuracy, and a final score of 0.769589.  The challenge
organizers notified our team that this result ranked first in the track.
\end{abstract}

\section{Introduction}
\label{sec:intro}

Referring video object segmentation predicts a binary mask sequence for the
object described by a language query.  MeViS emphasizes motion expressions,
which often cannot be resolved from a single frame~\cite{ding2023mevis,
ding2025mevis}.  The Audio track adds another source of uncertainty: the query
is spoken and must first be understood from audio.  Recognition errors can
change the object, its motion, or its number.  Moreover, some queries have no
matching target in the video, so a segmenter must sometimes return an empty
sequence rather than the most plausible visible object.

Earlier MeViS-Audio systems established a useful modular design.  APRVOS uses
speech recognition, visual existence checking, video segmentation, and mask
refinement~\cite{miao2026aprvos}.  ASR-SaSaSa2VA combines Qwen3-ASR,
SaSaSa2VA, and a separate no-target classifier~\cite{wang2026asr}.  VIRST-Audio
also separates transcription, segmentation, and existence prediction
~\cite{hong2026virstaudio}.  These systems show that speech recognition,
pixel-level tracking, and target existence are different decisions and should
not be forced into one output score.

Our system follows this modular view but reduces dependence on any single
video segmenter.  It creates several full-video mask tracks and selects the
one most consistent with the others.  The selection uses only predicted masks,
not validation or test labels.  Explicit motion and count rules are applied
only to queries for which the ordinary single-target assumption is
insufficient.  A final presence classifier handles no-target inputs while
protecting high-confidence structured predictions.

The main contributions are:
\begin{itemize}
  \item an audio-to-mask pipeline that combines strong open models without
        end-to-end retraining;
  \item a label-free, agreement-based rule for selecting a complete video mask
        track, followed by a small structured correction stage
  \item a video-level target-presence classifier designed to maintain high
        recall on valid targets while improving no-target accuracy.
\end{itemize}

\section{Challenge and Dataset}
\label{sec:challenge}

The 8th LSVOS Challenge at ECCV 2026 contains three tracks.  The
\textbf{MOSEv2 track} evaluates class-agnostic video object segmentation in
crowded scenes with occlusion, disappearance, small objects, and distractors;
it uses the MOSEv2 dataset~\cite{ding2025mosev2}.  The
\textbf{MeViS-Text track} evaluates referring video object segmentation from
written motion expressions.  The \textbf{MeViS-Audio track} uses spoken
versions of motion expressions and additionally evaluates whether the referred
target exists.  Both referring tracks are based on MeViS v2
~\cite{ding2025mevis}.

MeViS contains 2,006 videos, 8,171 annotated objects, and 33,072 motion
expressions in text and audio~\cite{ding2025mevis}.  Unlike referring
expressions dominated by color or object category, many MeViS queries describe
relative motion, temporal order, or interactions.  Correctly identifying the
target therefore requires evidence across time.  For the Audio track, the
input is a video $V=\{I_t\}_{t=1}^{T}$ and an audio query $A$.  The output is a
mask sequence $M=\{m_t\}_{t=1}^{T}$.  If the query has no matching target, all
$m_t$ must be empty.

\section{Method}
\label{sec:method}

\subsection{Overview}

\begin{figure}[t]
\centering
\includegraphics[width=\linewidth]{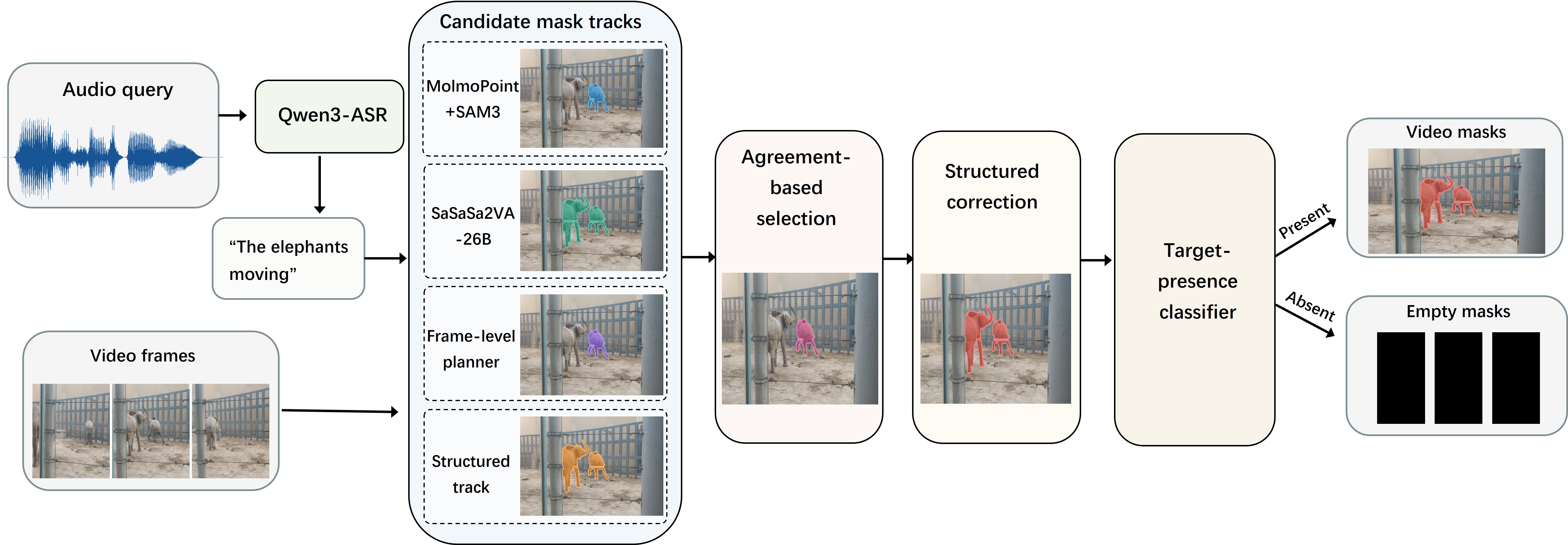}
\caption{System overview.  Audio is transcribed once, and all visual modules
receive the same text query.  Candidate tracks are selected by mask agreement;
the final presence decision either keeps the masks or returns an empty sequence.}
\label{fig:pipeline}
\end{figure}

Figure~\ref{fig:pipeline} shows the complete inference path.  The method first
transcribes the audio, then builds several candidate mask tracks.  An
agreement rule chooses one complete track per query.  Structured corrections
are applied to a small set of direction, count, and plural queries.  Finally,
the presence classifier either keeps the masks or replaces the whole sequence
with empty masks.

\subsection{Audio Transcription}

We use Qwen3-ASR-1.7B~\cite{shi2026qwen3asr} to convert the audio query into a
text expression
\begin{equation}
q = \operatorname{ASR}(A).
\end{equation}
Keeping transcription separate makes the visual stages easier to inspect and
allows all candidate generators to receive exactly the same expression.  The
ASR model produced transcripts for all 444 test queries.

\subsection{Candidate Mask Tracks}
\label{sec:candidates}

\paragraph{Point-guided SAM3 track.}
The main candidate uses MolmoPoint-8B~\cite{clark2026molmopoint} to predict
object points with timestamps from the video and transcript.  The video is
encoded at 6 frames per second and the prompt requests points at 1 frame per
second.  We use at most 128 input frames, with deterministic 96- and 64-frame
fallbacks when necessary.  These points initialize SAM3
~\cite{carion2025sam3}, which propagates masks through the original video.  A
parallel SAM2.1 propagation~\cite{ravi2024sam2} provides an independent
consistency signal for the frame-level fallback path.

\paragraph{Complementary tracks.}
We retain three candidates with different error patterns.  SaSaSa2VA-26B
~\cite{niu2025sasasa2va} directly predicts a text-conditioned video mask track
using its wrap-around-plus inference mode.  A frame-level planner chooses
between point-guided and fallback masks using point hits, agreement between
SAM2.1 and SAM3, support from neighboring frames, and mask similarity.  A
structured track is generated for queries that explicitly state horizontal
direction, object count, or plural subjects.  Qwen3-VL-30B-A3B
~\cite{bai2025qwen3vl} reads 16 sampled frames to extract these constraints,
and SAM3 forms the corresponding candidate tracks.

The point-guided route is available for 407 of the 444 test queries.  Queries
without a complete set of candidates use the full-coverage frame-level track.

\subsection{Agreement-Based Track Selection}
\label{sec:agreement}

For a query with $K$ candidate tracks, let $M_i^t$ be the mask from candidate
$i$ at frame $t$.  We first compute the mean temporal overlap between every
pair:
\begin{equation}
a(i,j) = \frac{1}{T}\sum_{t=1}^{T}
\operatorname{IoU}(M_i^t,M_j^t).
\label{eq:agreement}
\end{equation}
When both masks are empty, their frame-level overlap is defined as one.  The
score for candidate $i$ is its mean agreement with all other tracks,
\begin{equation}
s(i) = \frac{1}{K-1}\sum_{j\neq i}a(i,j), \qquad
i^* = \arg\max_i s(i).
\label{eq:selection}
\end{equation}
The output is the complete track $M_{i^*}$.  This is the medoid of the candidate
set under mask overlap, but in the remainder of the paper we simply call it
agreement-based selection.  It needs no ground-truth labels and avoids
calibrating confidence values from unrelated models.

On the test set, the selected source was point-guided SAM3 for 403 queries and
the structured candidate for two.  The frame-level and SaSaSa2VA candidates
were each selected once.  These counts cover the 407 queries with a full
candidate set.

\subsection{Structured Correction}
\label{sec:structure}

Ordinary track selection can preserve a visually plausible answer while still
missing a constraint such as ``moving left,'' ``two people,'' or a plural
subject.  We therefore apply a separate, conservative correction after
agreement selection.  A query is considered only if its transcript contains
an explicit direction, number, or plural construction.  Candidate tracks are
then checked for the requested horizontal motion and count; plural targets are
combined by mask union.  This stage changed 25 of the 444 test queries.  The
rules and thresholds were fixed before test inference.

\subsection{Target-Presence Classification}
\label{sec:presence}

The final stage predicts whether the transcript has a matching visual target.
It combines three scores: a Qwen3-VL-8B visual score, a direct Qwen2.5-Omni
audio-visual score~\cite{xu2025qwen25omni}, and the relative rank of the query
among all expressions for the same video.  For each score we compute seven
features: its raw value, within-video percentile, standard and robust
$z$-scores, gaps to the video-level maximum and minimum, and the logarithm of
the number of queries.  The resulting 21 features are passed to a balanced
logistic regression classifier.

The feature design and regularization were fixed on the training data.  For
the development domain, the classifier weights and threshold were refit with
video-grouped five-fold predictions.  The threshold was selected under a
minimum target recall of 0.97.  If the predicted probability is below the
threshold, all masks are replaced by empty masks.  As a safeguard, a query
accepted by the structured correction is not removed solely by the presence
classifier.  This safeguard restored three structured test predictions.  The
final classifier explicitly marked 42 test queries as no-target.

\section{Experiments}
\label{sec:experiments}

\subsection{Evaluation Protocol}

Mask quality is measured with region overlap $\J$ and boundary accuracy $\F$
following DAVIS~\cite{ponttuset2017davis}; $\J\&\F$ is their mean.  Target
accuracy (T-acc.) is the fraction of target-present queries kept by the
presence classifier, while no-target accuracy (N-acc.) is the fraction of
target-absent queries correctly rejected.  The challenge score is
\begin{equation}
\text{Final}=\frac{\J\&\F+\text{N-acc.}+\text{T-acc.}}{3}.
\end{equation}

Mask component studies use a frozen set of 100 target-present queries from the
official \texttt{valid\_u} split.  Presence results are reported as out-of-fold
(OOF) predictions with videos kept intact across folds.  No test masks or
per-query test labels were available when fixing the final rules.

\subsection{Implementation Details}

Qwen3-ASR-1.7B is used without task-specific fine-tuning.  MolmoPoint-8B runs
in 4-bit form, and Qwen3-VL-30B-A3B uses its released AWQ weights.  SAM3 and
SAM2.1 propagate masks at the original frame resolution.  The SaSaSa2VA
candidate is the 26B checkpoint.  Agreement selection uses the same four
candidate types and the same
rule on validation and test data.  The final archive was checked against the
sample submission and contains 28,415 PNG masks in 50 video directories.

\subsection{Mask Components}

Table~\ref{tab:mask_ablation} isolates the two decisions added after the main
point-guided SAM3 candidate.  Agreement-based selection improves both region
and boundary quality.  Structured correction brings a further 1.15-point
gain in $\J\&\F$, with the largest change in queries involving direction or
multiple objects.

\begin{table}[t]
\caption{Mask component study on 100 target-present \texttt{valid\_u} queries.
All values are percentages.}
\label{tab:mask_ablation}
\centering
\begin{tabular}{lccc}
\toprule
Method & $\J$ & $\F$ & $\J\&\F$ \\
\midrule
Point-guided SAM3 & 71.07 & 78.13 & 74.60 \\
+ agreement selection & 71.69 & 78.60 & 75.15 \\
+ structured correction & \textbf{73.00} & \textbf{79.60} & \textbf{76.30} \\
\bottomrule
\end{tabular}
\end{table}

Agreement selection chose the point-guided track for 91 of the 100 \texttt{valid\_u}
queries and the frame-level track for nine.  The rule nevertheless improves
the mean because it changes only complete tracks whose masks are better
supported by the other candidates.  This conservative behavior also explains
why the main point-guided track remains dominant on the test set.

\subsection{Presence Classification}

Table~\ref{tab:presence} reports video-grouped OOF accuracy.  The classifier
was first designed on the training split and then refit to \texttt{valid\_u}
without changing its inputs, feature transformations, regularization, or
target-recall constraint.  The lower \texttt{valid\_u} N-acc. indicates a
shift in absent-target expressions, while target recall remains above the
predefined 0.97 requirement.

\begin{table}[t]
\caption{Video-grouped out-of-fold target-presence results.  The model design
is fixed before the \texttt{valid\_u} refit.}
\label{tab:presence}
\centering
\setlength{\tabcolsep}{7pt}
\begin{tabular}{lcccc}
\toprule
Split & Queries & Videos & N-acc. & T-acc. \\
\midrule
Training OOF & 781 & 75 & 94.94 & 97.68 \\
\texttt{valid\_u} OOF & 907 & 50 & 81.58 & 97.58 \\
\bottomrule
\end{tabular}
\end{table}

\subsection{Official Test Result}

Table~\ref{tab:test} gives the score returned by the official server for the
submitted package (submission ID 865991).  The organizers subsequently
notified our team that the entry placed first in the MeViS-Audio track.  The
result also shows that existence prediction is a major part of this benchmark:
N-acc. and T-acc. contribute two thirds of the final score.

\begin{table}[ht]
\caption{Official hidden-test result of our final submission.}
\label{tab:test}
\centering
\begin{tabular}{lcccccc}
\toprule
Method & $\J$ & $\F$ & $\J\&\F$ & N-acc. & T-acc. & Final \\
\midrule
Ours & 0.5698 & 0.6205 & 0.5952 & 0.7931 & 0.9205 & \textbf{0.769589} \\
\bottomrule
\end{tabular}
\end{table}

\section{Discussion and Limitations}
\label{sec:discussion}

The experiments support two practical observations.  First, a strong spatial
pointing model followed by a video mask propagator is an effective main path,
but independent tracks still help identify occasional identity switches and
missed objects.  Comparing masks is more stable than comparing model-specific
confidence scores.  Second, presence classification should be handled at the
video-query level.  Within-video relative features are useful because many
queries refer to related objects in the same scene and raw multimodal scores
are not directly comparable across videos.

The method also has clear limitations.  Agreement can preserve a common error
when all candidates follow the same distractor.  The structured rules cover
only explicit direction, count, and plural expressions and do not perform
general language reasoning.  Small objects and long occlusions remain hard for
point-guided propagation.  Finally, the drop from \texttt{valid\_u} mask quality to
the hidden-test $\J\&\F$ shows that the two domains differ substantially.
Future work should replace hand-written structured cases with a trained
temporal grounding model and calibrate target presence on a larger variety of
no-target examples.

\section{Conclusion}

We presented the winning entry of the ECCV 2026 8th LSVOS MeViS-Audio track.
The system separates audio transcription, video mask generation, track
selection, structured correction, and target-presence classification.  Its
central mask rule selects the candidate that agrees most with the other full
tracks and requires no test labels.  This simple modular design achieved a
final official score of 0.769589 and can be reproduced from fixed model
checkpoints and deterministic post-processing rules.

\bibliographystyle{splncs04}
\bibliography{main}

\end{document}